\documentclass{IEEEtran}

\IEEEoverridecommandlockouts
\usepackage{amsmath,amsfonts,amssymb}
\usepackage{graphicx}
\usepackage[colorlinks=true, allcolors=blue]{hyperref}
\usepackage{cite}
\usepackage{enumerate}
\usepackage[linesnumbered,ruled,lined]{algorithm2e}
\usepackage{booktabs} 
\usepackage{amssymb}
\usepackage{bbding}
\usepackage{amsmath,amssymb,amsfonts}
\usepackage{algorithmic}
\usepackage{graphicx}
\usepackage{textcomp}
\usepackage{xcolor}
\usepackage{graphicx}
\usepackage{amssymb}

\makeatletter
\newcommand*\bigcdot{\mathpalette\bigcdot@{.5}}
\newcommand*\bigcdot@[2]{\mathbin{\vcenter{\hbox{\scalebox{#2}{$\m@th#1\bullet$}}}}}
\makeatother
\def\BibTeX{{\rm B\kern-.05em{\sc i\kern-.025em b}\kern-.08em
		T\kern-.1667em\lower.7ex\hbox{E}\kern-.125emX}}

\usepackage{stfloats}

\title{A Hierarchical Spatio-Temporal Graph Convolutional Neural Network for Anomaly Detection in Videos}

\author{Xianlin Zeng, Yalong Jiang, Wenrui Ding, Hongguang Li, Yafeng Hao, Zifeng Qiu
	
	\thanks{Manuscript received December 24, 2020; revised August 20, 2021; accepted December 6, 2021. This work was supported by pre-research project under Grant 62076019. (Corresponding authors: Yalong Jiang; Wenrui Ding).}
	
	\thanks{X. Zeng is with the School of Electrical and Information Engineering, Beihang University, Beijing 100191, China (e-mail: zengxianlin@buaa.edu.cn).}
	\thanks{Y. Jiang, W. Ding, and H. Li are with the Unmanned System Research Institute, Beihang University, Beijing 100191, China (e-mail: allenyljiang@buaa.edu.cn; ding@buaa.edu.cn; lihongguang@buaa.edu.cn). }
	\thanks{Y. Hao is with the 54th Research Institute of CETC, Shijiazhuang 050081, Hebei Province, China (e-mail: datoushr@126.com). }
	\thanks{Z. Qiu is with the Laboratory of Aerospace Information Applications of CETC, Shijiazhuang 050081, Hebei Province, China (e-mail: qzf93@qq.com). }
	}

\begin{document} 
\maketitle

\begin{abstract}
Deep learning models have been widely used for anomaly detection in surveillance videos. Typical models are equipped with the capability to reconstruct normal videos and evaluate the reconstruction errors on anomalous videos to indicate the extent of abnormalities. However, existing approaches suffer from two disadvantages. Firstly, they can only encode the movements of each identity independently, without considering the interactions among identities which may also indicate anomalies. Secondly, they leverage inflexible models whose structures are fixed under different scenes, this configuration disables the understanding of scenes. In this paper, we propose a Hierarchical Spatio-Temporal Graph Convolutional Neural Network (HSTGCNN) to address these problems, the HSTGCNN is composed of multiple branches that correspond to different levels of graph representations. High-level graph representations encode the trajectories of people and the interactions among multiple identities while low-level graph representations encode the local body postures of each person. Furthermore, we propose to weightedly combine multiple branches that are better at different scenes. An improvement over single-level graph representations is achieved in this way. An understanding of scenes is achieved and serves anomaly detection. High-level graph representations are assigned higher weights to encode moving speed and directions of people in low-resolution videos while low-level graph representations are assigned higher weights to encode human skeletons in high-resolution videos. Experimental results show that the proposed HSTGCNN significantly outperforms current state-of-the-art models on four benchmark datasets (UCSD Pedestrian, ShanghaiTech, CUHK Avenue and IITB-Corridor) by using much less learnable parameters.  
\end{abstract}


\section{Introduction}
\label{sec:intro}  

Human-related anomaly detection is the task of localizing from videos the activities that do not match regular patterns. The sophistication of anomaly semantics as well as the imbalance problem in anomaly-related datasets pose the need for detailed annotations that require expensive human labor. However, even detailed annotations can not ensure the generalization of real-world scenarios. It is, therefore, necessary to formalize the task as an unsupervised learning task that requires automatically discriminating small quantities of behavioral irregularity from the vast majority of normal events and is extremely challenging. This article concentrates on the study of abnormal events that are related to human behaviors and aims at developing models that characterize the feature patterns presenting in human behaviors and leverage the learned features to localize the short clips with behavioral irregularity from videos.

Existing unsupervised deep learning models develop the feature representations describing regular behaviors through training on video sequences with only normal events. The models reconstruct input data with learned embeddings which are optimized to produce low reconstruction errors on normal behaviors and the errors increase on behavioral irregularities. These works can be roughly divided into two categories: i) Reconstruction in the unit of pixels. The approach proposed in \cite{liu2018future} extracted features from historical frames with a Generative Adversarial Network (GAN) which then function in predicting future frames. Both temporal ranges and spatial locations of anomalies are determined by the GAN-based method. However, intensity-based features involve low-level clues which are quite noisy. For instance, the entanglement of background regions with foreground regions is difficult to be tackled. More seriously, real-world interferences such as dramatic changes in lightening conditions can lead to high reconstruction errors which are predicted by models as anomalies. As a result, the false alarm rate increases. On the other hand, the GAN-based models for per pixel reconstruction require huge computational burdens and are inefficient; ii) Reconstruction in the unit of body joints. These methods take advantage of the rich semantic features in critical regions and can describe human behaviors under lower noises and with higher efficiency. Paper \cite{morais2019learning} decomposed the movements of human skeletons into two sub-processes which describe the variation in moving speed and the variances in poses of each person, respectively, it proposed a model called Message-Passing Encoder-Decoder Recurrent Network (MPED-RNN) which is similar to the LSTM Auto Encoder (LSTM AE) proposed in \cite{srivastava2015unsupervised}. However, the RNN-based approaches are limited by the following disadvantages: First of all, the Seq2seq architecture proposed in \cite{martinez2017human} is fully-supervised and is only suitable for making predictions on the limited types of actions from training data. Secondly, this type of methods detect the abnormal behaviors of each person independently without fully considering the interactions among different people. Recently, Luo et al.\cite{luo2021normal} proposed Normal Graph for skeleton-based video anomaly detection, it explores the moving patterns of body joints in normal behaviors. Whereas, it only leverages skeleton-related features which are quite noisy in low-resolution videos. As a result, Normal Graph fails on low-resolution datasets, such as UCSD Pedestrian \cite{li2013anomaly}. By integrating high and low-level graph representations which describe the interactions among different people and show single person movements, respectively, the proposed approach achieves an improvement of about 7\% over Normal Graph and MPED-RNN with less learnable parameters on both high-resolution datasets such as ShanghaiTech and low-resolution datasets such as UCSD.

\begin{figure} [ht]
	\begin{center}
		\begin{tabular}{c} 
			\includegraphics[width=8.8cm]{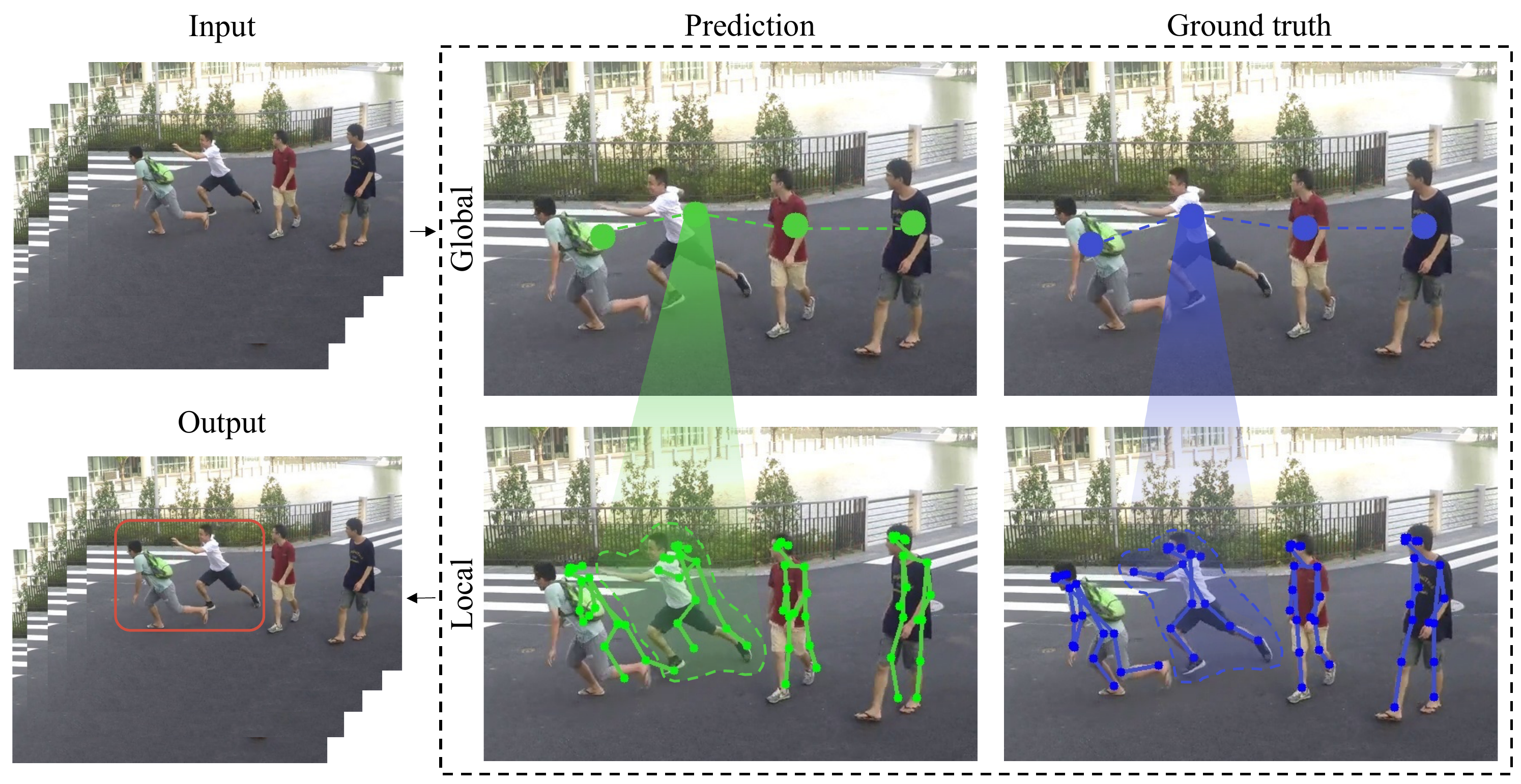}
		\end{tabular}
	\end{center}
	\caption[example] 
	{ \label{fig1} 
		The proposed scheme for hierarchical graph representations. The top row corresponds to the high-level graph representations where each node represents one individual. The bottom row shows low-level graph representations where each node describes one body joint. Individuals in the input video frame are encoded by high-level and low-level graph representations.}
\end{figure}

To leverage the strength of GCNN in analyzing structured data, we utilize STGCNN to effectively encode the spatial and temporal embeddings of human skeletons and determine the temporal inconsistency in human behaviors based on motion prediction. We name it Hierarchical Spatial-Temporal Graph Convolutional Neural Network (HSTGCNN). The HSTGCNN includes three components: a spatio-temporal graphical feature extractor, a future trajectory predictor, and an outlier arbiter. In detail, we firstly organize the inputs into a spatio-temporal graph whose nodes are human body joints from multiple frames and then feed it to the HSTGCNN to obtain intermediate feature representations that describe the motion vectors of semantic joints and indicate both directions and extents of movements. For instance, the action is more likely to be fighting if limbs are with much more strenuous movements than the head and torso. The HSTGCNN is trained on the input graph representations which correspond to normal activities and achieves the capability to accurately predict the trajectories of human joints in normal behaviors. The coordinates of human joints in historical frames, as well as future frames, are organized in the form of hierarchical graph representations, as is shown in Fig.~\ref{fig1}. A high-level graph representation is composed of nodes each of which represents a person, it characterizes the relative positions and interactions among individuals, the moving speed of each identity is also encoded in the graph representations. A low-level graph representation indicates the pose of a person and is leveraged in the detection of abnormal actions of a single individual.

Furthermore, the weighted combination of different levels of graph representations enables the proposed framework to exhibit robustness to the variances in scenes. Different branches conduct inference on different levels of graph representations and are better at handling different scenes some of which are with dense small human objects while others contain sparse and large humans. To acquire an understanding of scenes and determine the weights of branches accordingly, we propose to use optical flow fields and average sizes of human bounding boxes and skeletons to cluster videos into different groups that correspond to different scenes. We determine the weight of branches which minimizes the loss of predictions in normal behaviors. For scenes with dense crowds and small people, higher weights are assigned to high-level graph representations while in sparse scenes, the weights of low-level graph representations are increased. Finally, the outlier arbiter weightedly sums the predictions from independent branches to obtain the final anomaly score. The model can produce effective detections of abnormal events only by being trained in an end-to-end manner on a small number of video sequences without anomalies. Experimental results on standard benchmarks, including UCSD Pedestrian \cite{li2013anomaly}, ShanghaiTech \cite{luo2017revisit}, CUHK Avenue \cite{lu2013abnormal} and IITB-Corridor \cite{rodrigues2020multi}, demonstrate the effectiveness and efficiency of our approach, it outperforms state-of-the-art approaches while using much less learnable parameters.

The main contributions of this article can be summarized as follows: 

(1) We propose a Spatio-Temporal Graph Convolutional Architecture that integrates branches each of which corresponds to the graph representations in a certain level. Low-level graph structures represent the spatial and temporal embeddings of human body joints. High-level graph structures not only characterize the speed and direction of each pedestrian but also represent behavioral irregularities by modeling the interactions among multiple identities.

(2) Different branches are better at different scenes (sparsely distributed people with high resolutions and dense small-scale humans) grouped by clustering. The predictions from different branches are weightedly combined to obtain the final anomaly score and achieve an understanding of scenes.

(3) Compared with other STGCNN-based methods, our proposed framework can overcome drastic changes in scenes by using high-level graph representations combined with skeleton-based features and perform well on both high-quality videos and low-quality ones where humans are of low resolutions.

The rest of the paper is organized as follows. Section II describes the related work on human pose estimation, human action recognition, and video anomaly detection. In Section III, the pre-processing of human skeletons is given full description firstly. Then the details of the overall framework and each part are presented. Section IV first introduces dataset and implementation details, then conducts several comprehensive experiments to demonstrate the effectiveness of the proposed model and perform analysis on errors. Finally, Section V summarized.

\section{Related Work}

 \subsection{Human pose estimation in videos}
 
 Human pose estimation in computer vision is the task of estimating the pose of articulated bodies. Existing methods for pose estimation are divided into 2D approaches and 3D ones. Among the 2D pose estimation methods, the first type of method only works on images with a single person \cite{newell2016stacked}. These methods firstly localize semantic parts and then leverage the connections between parts to describe gestures. Single-person pose estimations can not be applied to many real-world scenarios. The second type of method conducts multi-person pose estimation by using a human detector to localize humans before estimating the pose of each person. This type of methods are named top-down methods and include Mask-RCNN \cite{he2017mask}, RMPE \cite{fang2017rmpe}, CPN \cite{chen2018cascaded}, etc. The third type of method detects the body joints of all people in an image before grouping them into people. This type of methods are called bottom-up methods and include Open Pose \cite{cao2018openpose}, Deep Cut \cite{pishchulin2016deepcut}, Deeper Cut \cite{insafutdinov2016deepercut} etc.

 \subsection{Human action recognition in videos}
 
 Related studies have conducted human action recognition based on pose estimations. Earlier work used simple geometric models in representing the structures of human bodies (such as a two-dimensional contour model \cite{fathi2008action} and a three-dimensional cylinder model \cite{zouba2008monitoring}), and focused on the dynamics of external contours. Recently, some methods have been proposed to use interactive recurrent networks for modeling human motions on hard scenarios \cite{alahi2016social,gupta2018social}. The above-mentioned approaches all modeled each human as one rigid object, ignoring the local movements of human body joints. To pay attention on both moving speed and pose variances of people, the approaches \cite{fragkiadaki2015recurrent,villegas2017learning} took human skeletons as the input to RNN and modeled human movements in this way. Recently, paper \cite{du2015hierarchical} proposed to divide human skeletons into five parts and feed them into five independent RNNs for feature extraction. The model has been successfully applied in some cases \cite{zheng2018unsupervised,liu2018multi,liu2016spatio}. However, the RNN-based models suffer from error accumulation, gradient explosion, and gradient vanishing problems due to complex structures. The problem is avoided in our approach with a shallow structure. These methods can only detect some fixed types of actions pre-defined in training data and are difficult to generalize to unpredictable abnormal behaviors.
 
 Skeleton based action recognition is an emerging topic. Skeleton and joint trajectories of human bodies are more robust to illumination changes and scene variations than pixel-level features. As a result, it has been widely used in video action recognition tasks \cite{shi2019two,zhu2016co}, pose tracking \cite{gall2009motion}, etc. \cite{zeng2019graph} applied the GCN over the graph to establish the relationship between two action proposals to boost the performance of temporal action localization. It achieved good generalization ability and the method \cite{mohamed2020social} also received great attention. \cite{wang2020learning} also built the GCN on the individual level for action recognition but lacked the modeling of interactions.
 
 However, the above-mentioned models for action recognition suffer from the following problems: i) Most of them \cite{gu2018ava} take into account the spatial and temporal relations but only work on temporally trimmed videos which include only one fixed type of action across an entire video. ii) The reasons leading to anomalies are diverse and unpredictable, resulting in sophisticated semantics that can hardly be covered by existing training data. Existing supervised action recognition methods only detect a limited number of action types that are annotated in training data \cite{wang2020learning}. Inspired by the application of unsupervised domain in pedestrian re-identification \cite{zhao2020unsupervised}, our proposed approach can detect abnormal behaviors in untrimmed videos through unsupervised training and does not require the types of anomalies to appear in training data.

 \subsection{Human anomaly detection in videos}
 
 In the community of video processing and computer vision, early algorithms treated each frame as an independent sample, superpixel techniques were employed to figure out the motion orientations of objects and conducted anomaly detection simultaneously \cite{yuan2016anomaly}. Paper \cite{mohammadi2021image} conducted a detailed investigation into the videos and images deep learning based anomaly detection methods. Typical machine learning techniques involve clustering ($K$-means \cite{lee2015motion}, GMM \cite{leyva2018fast}, OC-SVM \cite{yin2008sensor}, etc.) and reconstructing discriminant analysis (such as PCA \cite{candes2010robust,xiong2011direct,debruyne2010robust} and SC \cite{liu2018future,zhao2011online,cong2011sparse,zhu2014sparse,yuan2018structured}, etc.). When dealing with large-scale data with complex abnormal patterns, these methods often fail to achieve desired results.
 
 The advancements of deep learning have brought many advantages over traditional methods in feature engineering. However, the determination of anomalies is still based on hand-crafted approaches. Typical deep learning-based feature extractors include Auto Encoder (AE) \cite{chong2017abnormal}, target detection neural network \cite{hinami2017joint}, 3D Convolutional Network (C3D) \cite{chu2018sparse} and so on. Although the CNN-based method has achieved great success in many visual tasks, the applications of CNN in anomaly detection are still limited by two constraints. Firstly, CNN shows a heavy reliance on a large amount of training data with full supervision. Secondly, CNN can not deal with non-euclidean graphical data where the number of neighbors around each node is not fixed, as is often the case in human-related graph representations.
 
 As more research efforts been invested into anomaly detection, recent studies combined the feature extraction step with the model training step and proposed deep learning methods in an end-to-end manner, such as VAE \cite{wang2018generative}, Generative Adversarial Network (GAN) \cite{sabokrou2018adversarially}, Recurrent Neural Network (RNN) \cite{luo2017revisit} and Long Short-Term Memory (LSTM) \cite{lee2018stan}, etc. Among them, paper \cite{ravanbakhsh2017abnormal} used a U-Net-based model to achieve the cross-modal reconstruction of video frames based on generative approaches and optical flow estimations, the reconstruction error was used to determine anomalies. The network utilized long skip connections to reduce the information loss caused by the reduction in dimensionality and achieved better results than AE. Besides, the GAN-based network improved the performance of the generator through adversarial training. For example, the research \cite{zenati2018efficient} used Bidirectional GAN (BiGAN) to perform anomaly detection, pixel intensity loss and discriminator loss were also leveraged. Paper \cite{morais2019learning} proposed a Message-Passing Encoder-Decoder Recurrent Network (MPED-RNN) with human skeletons as inputs, it suppressed pixel-level noises which commonly exist in optical flow estimations, the interpretability of abnormal semantics was also improved. However, using recursive architecture to extract skeleton features is not the optimal solution in that error accumulating in sequential predictions made by RNNs and too many learnable parameters lead to the consumption of huge memory spaces. Paper \cite{luo2021normal} proposed a single-level graph structure called Normal Graph for skeleton-based video anomaly detection, which takes body joint locations of each pedestrian as inputs and simply calculates the mean square error between the predicted joints and the ground truth. This method shows a heavy reliance on skeleton estimations which are negatively influenced by the reduction in resolutions of human objects. The high-level graph representations in the proposed framework address this issue. Besides, Normal Graph lacks a description of the interactions among multiple people, as result, it suffers from poor accuracy. 
  
 Unlike the above-mentioned works, inspired by the success of graph structures in the question answering system \cite{huang2020location} which was based on dynamic image understanding and in the traffic forecast system \cite{yu2017spatio}, we propose a Hierarchical Spatial-Temporal Graph Convolutional Neural Network (HSTGCNN) which encodes the spatial and temporal relations while constructing hierarchical graph representations in describing group behaviors in anomaly detection. The model can characterize not only the interactions among multiple individuals but also the local movements of each individual by Spatial-Temporal characteristics of graph representations in various resolution video frames and achieve a comprehensive understanding of scenes.

\section{Approach}

Human-related anomalies can be detected with unsupervised methods through predicting human motions in future video frames based on historical trajectories. The models are trained to minimize the errors between the predictions and the ground truth only on normal behaviors. Since spatio-temporal graph convolutional networks can comprehensively represent the structures of human bodies and dynamics, we propose a hierarchical graph representation in jointly modeling the inter-connections among identities and the correlations among semantic parts within the same identity. Meanwhile, the weighted combination of multiple branches that are better at handling different scenes is proposed to dynamically adjust the overall model for anomaly detection under various scenes. The feature representations of scenes, including the densities and scales of identities, are proposed to determine the weights of branches. In this section, we will describe the proposed framework in detail.

 \subsection{Problem formulation and skeleton modeling}
 
 Denote the set of human skeleton trajectories in $T$ consecutive video frames as $S$, expressed as:
 
 \begin{equation}
 \label{eq1}
 S=\left\{f_{t} \mid t \in\{1, \cdots, T\}\right\} \, ,
 \end{equation}
 
\noindent where,
 
  \begin{equation}
  \begin{aligned}
 \label{eq2}
 f = &\left\{\left(x_{m, n}, y_{m, n}\right)\right\}  \\
  & \mid m \in\{1, \cdots, M\} ; n \in\{1, \cdots, N\} \, ,
  \end{aligned}
 \end{equation}
 
\noindent where $m$ is the index of individuals in a single frame, $n$ is the index of each individual's joints, and $f_{t}$ represents the set of coordinates in the $t$-th frame.
 
 \begin{figure} [ht]
 	\begin{center}
 		\begin{tabular}{c} 
 			\includegraphics[width=8.8cm]{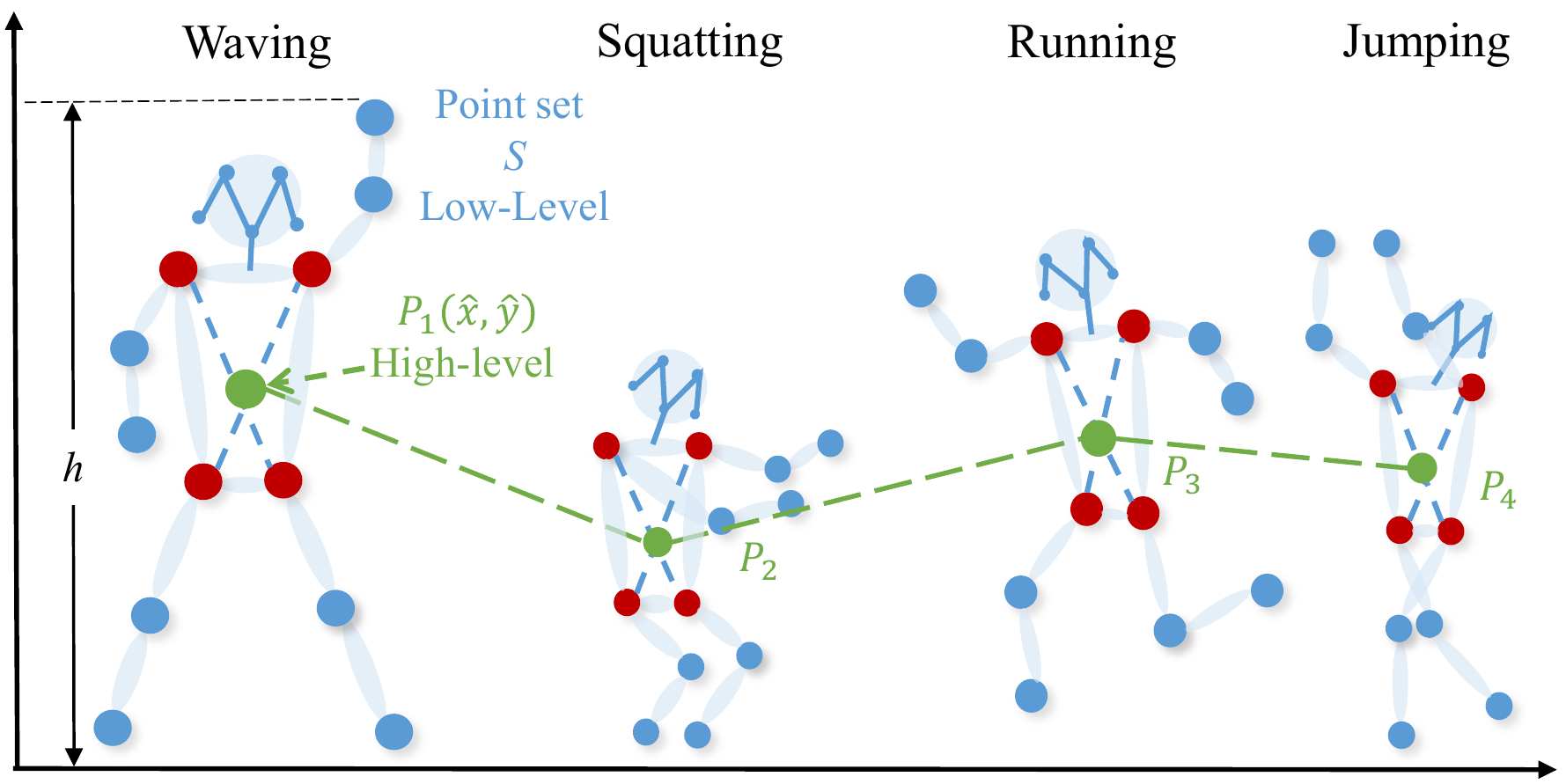}
 		\end{tabular}
 	\end{center}
 	\caption[example] 
 	{ \label{fig2} 
 		The demonstration of the high/low-level graph representations. Low-level graph representations are the blue graphs denoted by $S$ and each node corresponds to a body joint. High-level graph representations are shown by the green graph where each node is denoted by $P_{i}$, each node represents the feature embeddings of one identity. $S$ denotes the set of all human skeleton trajectories in $T$ consecutive video frames; $P_{i}$ is the geometric center of the $i$-th person.}
 \end{figure} 
 
 The input data denoted by $S$ is organized in a graph structure and is fed to hierarchical graph convolutional networks which describe the motion of single-individuals and the correlations among multiple individuals. The proposed graph convolutional network for anomaly detection learns to predict the future trajectories of each individual, including both the trajectories of human centers and those of body joints. For video sequences with only normal behaviors, the model can accurately predict the trajectories of joints in future frames because the motions in all short clips subject to the same distribution. For the videos with anomalies, there is an obvious difference between joint trajectories and the model's predictions in the occurrences of anomalies. Typical anomalies involve dramatic changes in posture and rapid movements. In order to comprehensively describe the difference between predicted graph structures and the ground truth, we combine high-level graph structures with low-level graph structures, the former regards human bodies as rigid objects while the latter treats different body joints as different nodes. 
 
 Some pre-processing methods are required to reduce the influences of task-irrelevant factors on anomaly detection. The most prominent factor is the variances in human sizes, the variances in larger humans' motions are obviously larger and this is the same with prediction errors, we propose to conduct normalization on joint coordinates with respect to human sizes while maintaining geometric centers. The human skeleton before pre-processing is shown in Eq. \ref{eq2}. The coordinates are normalized with respect to the heights of bounding boxes which are the maximal difference in vertical coordinates of joints' vertical locations: $h_{m}=\max \left(y_{m,n}\right)-\min \left(y_{m,n}\right)$. Upon normalization, the movement of each body joint is decomposed into the movement of the overall body and the relative motion of the joint with respect to its body center. The body geometric center $P$ is obtained by computing the mean value of the coordinates of the four joints from the human torso, as shown in Eq. \ref{eq3} and \ref{eq4}:

 \begin{equation}
 \label{eq3}
 P=\left(\hat{x}_{m}, \hat{y}_{m}\right)\, ,
 \end{equation}
 
 \begin{equation}
 \label{eq4}
 {{\hat{x}}_{m}}=\frac{\sum\limits_{n\in k}{{{x}_{m,n}}}}{4},{{\hat{y}}_{m}}=\frac{\sum\limits_{n\in k}{{{y}_{m,n}}}}{4},k\in [5,6,11,12]\, ,
 \end{equation}
 
\noindent where $k$ represents the indices of the four selected points among $17$ body joints, which represent different human body parts, the set of joints included in $k$ is shown by the four red points in Fig.~\ref{fig2}. More details about the definitions of the 17 body joints are provided in HRNet \cite{sun2019deep}.
 
 First of all, we collect all human skeletons in each frame, and calculate the geometric centers $P_{i}$ of the $i$-th person according to Eq. \ref{eq3} and \ref{eq4} to obtain high-level graph nodes. Then, we construct the adjacency matrix according to Eq. \ref{eq10} to obtain high-level graph edges. High-level graph representations are denoted as $f_{m}^{g}$, which encodes the relative position of all individuals,

 \begin{equation}
 \label{eq5}
 f_{m}^{g}=\left\{\left(\hat{x}_{m}, \hat{y}_{m}\right) \mid m \in\{1, \cdots, M\}\right\}\, .
 \end{equation}

 By subtracting the coordinates of body joints by the geometric centers in $P$, we can focus on the variances in poses. We construct low-level graph nodes encoding the proposal relative positions of body joints with respect to body center. The low-level graph edges between semantically connected body joints are 1 and others 0. Low-level graph representations are denoted as $f_{m,n}^{l}$,
 
 \begin{equation}
 \begin{aligned}
 \label{eq6}
 f_{m,n}^{l}=& \left\{\left(\tilde{x}_{m,n}, \tilde{y}_{m,n}\right)\right\}  \\
  & \mid m \in\{1, \cdots, M\} ; n \in\{1, \cdots, N\} \, ,
  \end{aligned}
 \end{equation}

\noindent where,
 
 \begin{equation}
 \label{eq7}
 \tilde{x}_{m,n}=\frac{x_{m,n}}{\frac{h_{m}}{2}}, \tilde{y}_{m,n}=\frac{y_{m,n}}{h_{m}}\, . 
 \end{equation}

 Fig.~\ref{fig2}. shows the construction of the high/low-level graph representations, the coordinate of each body joint can be expressed as the summation of a global coordinate which denotes the center of the individuals, and a local coordinate which denotes the offset of the joint from the body center. High-level graph representations encode the speed and relative positions of different individuals. The anomalies caused by the outliers in pedestrian movement speed or suddenly dispersed crowds can be detected by high-level graph representations. Low-level graph representations are better at handling posture changes, such as fighting, falling, and other irregular behaviors. Abnormal events are usually contributed by a variety of complex factors and we leverage three prediction branches that are better at expressing and detecting different types of anomalies by concentrating on different levels of graph representations. By adjusting the weights of different branches, we can adjust the overall framework to fit different scenarios and improve the robustness of the method.
 
 \subsection{Network architecture}
 
 \subsubsection{Overall network}
 
\begin{figure*} [ht]
 	\begin{center}
 		\begin{tabular}{c} 
 			\includegraphics[width=18.1cm]{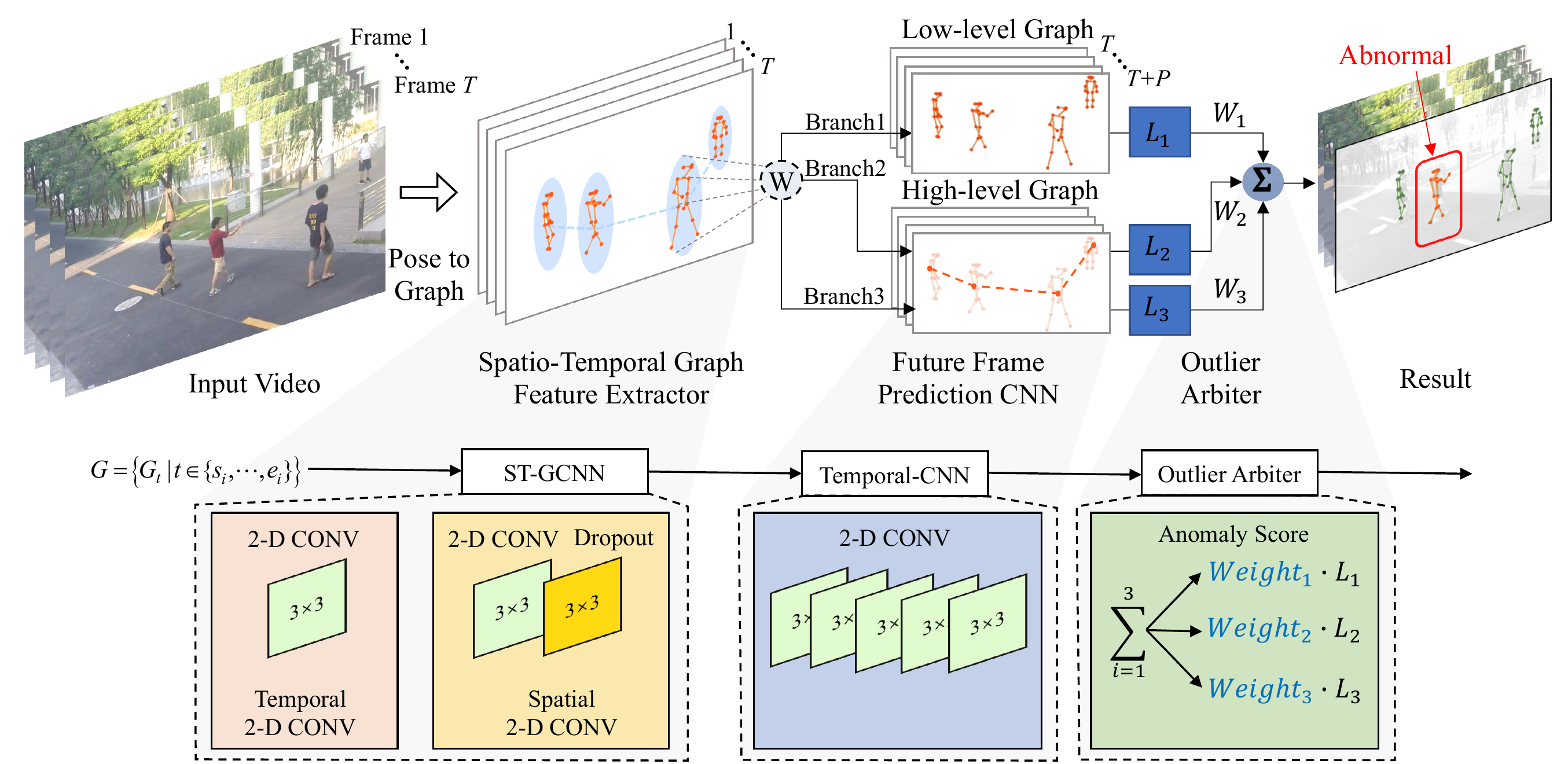}
 		\end{tabular}
 	\end{center}
 	\caption[example] 
 	{ \label{fig3} 
 		Structure of the proposed framework. The HSTGCNN model consists of three major parts: a spatio-temporal graphical feature extractor (STGCNN), a future frame predictor (Future Trajectory Prediction CNN), and an outlier arbiter (Anomaly Score).}
 \end{figure*}

 The HSTGCNN model proposed in this paper consists of three major parts: a spatio-temporal graphical feature extractor, a future frame predictor, and an outlier arbiter. The spatio-temporal graphical feature extractor consists of a spatio-temporal graph convolutional neural network and performs spatio-temporal convolutional operations on the graph representations of all skeletons in the historical frames to extract features. The future frame predictor consists of a temporal convolutional network. Taking the graph representations of the human skeletons as input, the temporal convolutional network is expected to predict the future skeleton trajectories with convolutional operations. The outlier arbiter firstly feeds the outputs from the second part into multiple branches, and then weightedly sum up the predictions from all branches to obtain the anomaly score. Fig.~\ref{fig3}. shows the structure of the model.
 
 \subsubsection{Local network structure}
 
   \begin{itemize}
 	\item Construction of graph representations 
 	
 	In Section III-A, we used the $xOy$ coordinate system to represent the set $S$ of all human skeleton trajectories in a video with t frames and then convert it into a graph structure as the input of the STGCNN. Firstly, construct a set of spatial graphs $G_{t}$ to represent the graph structure of $S$. $G_{t}$ is defined as:
 	
 	 \begin{equation}
 	\label{eq8}
 	G_{t}=\left(V(G_{t}), E(G_{t})\right)\, ,
 	\end{equation}
 	
 	where $V^{g}(G_{t})=\left\{V_{i}(G_{t}) \mid \forall i \in\{1, \cdots, M\}\right\}$ and $V^{l}(G_{t})=\left\{V_{i}(G_{t}) \mid \forall i \in\{1, \cdots, N\}\right\}$ denote two sets of vertices in a high-level graph and a low-level graph, respectively. The coordinates of geometric centers $P$ and body joints $f$ compose $V^{g}(G_{t})$ and $V^{l}(G_{t})$, respectively. 
 	
 	$E(G_{t})$ is a set of edges in the graph $G_{t}$, expressed as:
 	
 	\begin{equation}
 	\label{eq9}
 	E(G_{t})=\left\{e_{t}^{i j} \mid \forall i, j \in\mathbb{N^{*}}\right\}\, .
 	\end{equation}
 	
 	If $V_{i}(G_{t})$ and $V_{j}(G_{t})$ are connected, $e_{t}^{i j}=1$, otherwise, $e_{t}^{i j}=0$. Besides, an adjacency matrix indicating the relative positions of nodes is integrated in the input to HSTGCNN. HSTGCNN uses $a_{t}^{i j}$ to represent the distance between the $i$-th node and the $j$-th node. The value is determined by the following formula:
 	
 	\begin{equation}
 	\label{eq10}
 	a_{t}^{i j}=\left\{\begin{array}{cc}
 	\!\!\!\frac{1}{\left(V_{i}(G_{t})-V_{j}(G_{t})\right)^{2}} &\scriptstyle \!\!\!\!,\hspace{1ex} \left(V_{i}(G_{t})-V_{j}(G_{t})\right)^{2} \neq 0 \\
 	\scriptstyle \!\!\!0  &\scriptstyle \!\!\!\!,\hspace{1ex} \left(V_{i}(G_{t})-V_{j}(G_{t})\right)^{2}=0
 	\end{array}\right.\, \!\!\!,
 	\end{equation}
 	
 	where $V^{g}(G_{t})=\left\{V_{i j}(G_{t}) \mid \forall i, j \in\{1, \cdots, M\}\right\}$ and $V^{l}(G_{t})=\left\{V_{i j}(G_{t}) \mid \forall i, j \in\{1, \cdots, N\}\right\}$ correspond to high-level graph edges and low-level graph edges respectively, we calculate each $e_{t}^{i j}$ to construct a weighted adjacency matrix $A$.

 	\item STGCNN model layout 
 	
 	In order to facilitate better convergence of the model, we normalize the adjacency matrix. The adjacency matrix $A$ of $T$ frames can be expressed as a set: $A_{t}=\left\{A_{1}, \cdots, A_{T}\right\}$, we symmetrically normalize each $A_{t}$, according to Eq. \ref{eq11}:
 	
 	\begin{equation}
 	\label{eq11}
 	A_{t}=D^{-\frac{1}{2}} \hat{A}_{t} D^{-\frac{1}{2}}\, ,
 	\end{equation}
 	
 	where $D$ is the diagonal nodal degree matrix of $\hat{A}_{t}$. The operation in the $(l+1)$-th layer can be expressed by the formula:
 	
 	\begin{equation}
 	\label{eq12}
 	H^{(l+1)}=\sigma\left(D^{-\frac{1}{2}} A D^{-\frac{1}{2}} H^{(l)} W^{(l)}\right)\, ,
 	\end{equation}
 	
 	where $W^{(l)}$ is the matrix of trainable parameters at layer $l$ and $\sigma$ is the nonlinear activation function.
 	
 	\item Future Frame Prediction CNN model layout
 	
 	STGCNN extracts spatio-temporal graph representations that serve the purpose of predicting future trajectories. Future Frame Prediction CNN operates directly on the temporal dimension of graph embeddings $V_{t}$ and leverages the temporal clues for prediction. It is composed of five residual convolutional layers, it takes in four frames of input and outputs one frame with predicted human poses.
 	 		
 \end{itemize}

Using the spatio-temporal graphical feature extractor, we get the feature representations in the form of 4-dimensional tensors: $(f, t, m, n)$, as detailed in Eq. \ref{eq1} \& \ref{eq2}. We exchange the first and second dimensions of the tensor to produce the input of the future frame predictor. The future frame predictor uses the residual convolutional network \cite{he2016deep} as the backbone and predicts the coordinates of future samples. The difference between predicted trajectories and ground truth is the loss for training the model. The outlier arbiter does not require training, it is only responsible for calculating the anomaly score to judge the level of anomalies as will be discussed later in Eq. \ref{eq14} to \ref{eq17}.

\subsection{Model training}

 \subsubsection{Fixed number of frames as input}

Based on the above-mentioned approaches, we propose to track multiple individuals in multiple video frames. However, HSTGCNN needs to be trained on a sequence with a fixed length of $T$. We use a sliding window strategy to extract video clips with $T=5$ frames. The reason for not selecting a too large $T$ is due to some limitations. For instance, due to occlusions and other problems, long trajectories may contain inconsistencies in human appearances or movements. The determination of $T$ is with reference to \cite{liu2018future}. An appropriate $T$ ensures stable training. During training, we propose to use the human skeletons in four consecutive video frames as the input of the reconstruction model to predict all human skeletons in the fifth video frame.

 \subsubsection{Model configuration}
 
 This section focuses on the spatio-temporal graphical feature extractor and the future frame predictor. We use PReLU \cite{7410480} as the activation function $\sigma$ in all layers. According to Mohamed et al.'s \cite{mohamed2020social} research, when the number of STGCNN layers increases, the performance decreases, so spatio-temporal graphical feature extractor includes one STGCNN layer and the future frame predictor has five Temporal-CNN layers.
 
 \subsubsection{Loss function}
 
 In order to better reconstruct the trajectory of individuals in normal mode in unsupervised learning, we choose the mean square error (MSE) loss function to calculate the loss between the model output and the ground truth. Such as the formula:
 
 \begin{equation}
 \label{eq13}
 M S E=\frac{1}{\mathrm{T}} \sum_{t=1}^{T}\left(\hat{f}_{t}^{g+l}-f_{t}^{g+l}\right)^{2}\, ,
 \end{equation}
 
\noindent where $\hat{f}^{g+l}$ represents the global and local locations of prediction and $f_{t}^{g+l}$ represents the ground truth.

\subsection{Video clustering}
Under different scenarios, different strategies are adopted with different weights of branches. In this work, we divide training videos into different groups and use clustering methods to group videos with similar scenes. In implementations, we collect the number of individuals, average sizes of human bounding boxes and skeletons, optical flow fields and other information as the features of the scenarios in the video as the input of the $K$-means for initialization. Next, we determine a set of appropriate weight coefficients, $(W_{1}, W_{2}, W_{3})$, so as to minimize the loss of weightedly summing the three prediction branches in each group on the training set. For instance, for scenes with dense crowds, higher weights are assigned to high-level graph representations while in sparse scenes, the weights of low-level graph representations are increased.
 \subsection{Anomaly detection}
 
 In order to calculate the anomaly score of each frame in the video of the test dataset, we propose the third part of the model: the outlier arbiter. Based on the hierarchical graph representation inference introduced in Section III-B, three independent branches are built to provide predictions on the level of anomalies. The outlier arbiter combines the branches to obtain the anomaly scores through the weighted summation, the process is divided into the following several steps:

 \begin{enumerate}[1.]
 	\item In the same way as the training set, we take a sliding window with a stride of 1 and a window size of 5 frames on the test set. Every 4 frames are leveraged by the predictor to predict the fifth frame.
 	
 	\item $L_{1}$ denotes the prediction error on the poses of each independent individual, which is computed with low-level graph representations in branch 1, according to Eq. \ref{eq6} \& \ref{eq7}:
 	
 	\begin{equation}
 	\label{eq14}
 	L_{1}=\frac{1}{T_{\mathrm{e}}-T_{s}} \sum_{t=T_{s}}^{T_{e}}\left(\hat{f}_{t}^{l}-f_{t}^{l}\right)^{2}\, .
 	\end{equation}
 	
 	$L_{2}$ is the error in jointly predicting the center points of multiple people. It is computed with high-level graph representations in branch 2, according to Eq. \ref{eq5}:
 	
 	\begin{equation}
 	\label{eq15}
 	L_{2}=\max \left[\left(\hat{f}_{t}^{g}-f_{t}^{g}\right)^{2} \mid t \in\left\{T_{s}, \cdots, T_{e}\right\}\right]\, .
 	\end{equation}
 	
 	$L_{3}$ is the prediction error on the motion vectors of multiple people, it is computed by high-level graph representations in branch 3, according to Eq. \ref{eq5}:

 	\begin{equation}
 	\begin{aligned}
 	\label{eq16}
 	L_{3}=\max \bigg[ \left( \hat{f}_{t_{1}}^{g} \right. - & \left. \hat{f}_{t_{2}}^{g}\right)^{2} -  \Big(f_{t_{1}}^{g}- f_{t_{2}}^{g}\Big)^{2}\bigg]\\  
 	 & t_{1}, t_{2} \in\left\{T_{s}, \cdots, T_{e}\right\} ; t_{1} \neq t_{2}\, .
 	\end{aligned}
 	\end{equation}
 	
 	Obviously, if there is only one individual in the frame, the score $L_{3}$ is $0$. $\hat{f}_{t}$ represents the predicted locations and ${f}_{t}$ represents the ground truth in Eq. \ref{eq14} to \ref{eq16}.

 	\item Based on the calculation in step 2, assuming a video frame has a duration of $T$, We obtain anomaly scores on $T-4$ frames, the score on each frame is the summation of 3 terms. Here, the three branches independently characterize anomalies from different point-of-views. Finally, after assigning the same set of weights to each set of videos grouped by clustering, the overall anomaly score is obtained as is demonstrated in Eq. \ref{eq17}:
 	
 	\begin{equation}
 	\label{eq17}
 	L=W_{1} \bigcdot L_{1}+W_{2} \bigcdot L_{2}+W_{3} \bigcdot L_{3}\, ,
 	\end{equation}
 	
 	where $W$ represents the individual weight coefficient of each branch.
 	
 \end{enumerate}

We use Algorithm 1 to specifically describe the anomaly score calculation process of the above algorithm.

\begin{algorithm}[ht]
	
	\SetKwInOut{Majorization}{Majorization}\SetKwInOut{Minimization}{Minimization}
	\SetKwData{set}{set}
	\SetKwInOut{Initialization}{Initialization}
	\vspace{2mm}
	\KwIn  {The high-level graph representations ${f_{t}}^{g}$, the low-level graph representations ${f_{t}}^{l}$, the weights of branches $W$.}
	\KwOut {Final anomaly score $L$.}

	\vspace{2mm}
	Initialize ${f_{0}}^{g}$,${f_{0}}^{l}$ and others\;
	\For{ $T_{s} = 0 \sim T - 5$ }{
		$T_{e}=T_{s}+5$\;
		
		\For{all $t \in \{ T_{s},\cdots ,T_{e} \}$ } {
			Calculate $L_{1}$, $L_{2}$, $L_{3}$ with Eq. \ref{eq14}, \ref{eq15}, and \ref{eq16}\;
			Minimize $W_{1} \bigcdot L_{1} + W_{2} \bigcdot L_{2} + W_{3} \bigcdot L_{3}$ with respect to $W_{1}$, $W_{2}$, $W_{3}$ on training set\;
			Share the weights of branches $W$ in the same group of videos\;
			Calculate anomaly score $L$ with Eq. \ref{eq17}\;
		}
	}
	\caption{\small{ Calculate anomaly score.}}
	\label{alg}
\end{algorithm}

\begin{figure*} [ht]
	\begin{center}
		\begin{tabular}{c} 
			\includegraphics[width=18.1cm]{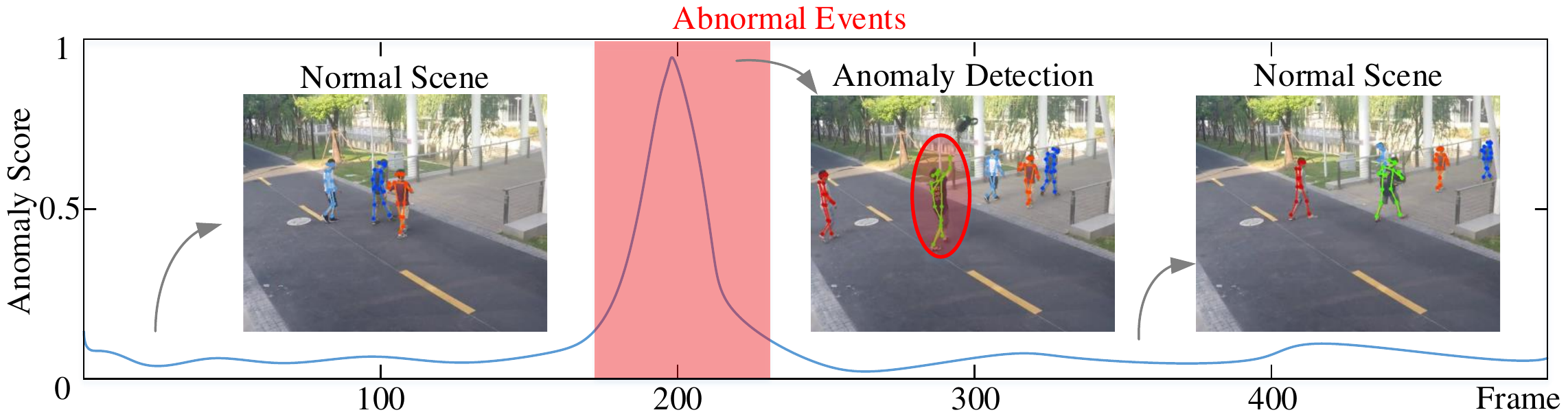}
		\end{tabular}
	\end{center}
	\caption[example] 
	{ \label{fig4} 
		Anomaly score graph for a testing video in the HR-ShanghaiTech dataset. The peak in anomaly scores corresponds to the frame with anomalies which are marked by red circles indicate frame-level ground truth of anomaly score.}
\end{figure*}

\section{Experiment}
 \subsection{Datasets and pre-processing}
 \subsubsection{Datasets}
 
 To demonstrate the effectiveness of the proposed method, we conduct experiments on four public datasets: the UCSD Pedestrian dataset \cite{li2013anomaly}, the ShanghaiTech Campus dataset \cite{luo2017revisit}, the CUHK Avenue dataset \cite{lu2013abnormal} and the IITB-Corridor dataset \cite{rodrigues2020multi}. The training set of these data sets contains only normal events, while the test set contains normal and abnormal events.
 
  \begin{itemize}
  	\item UCSD Pedestrian dataset is acquired with a stationary camera mounted at an elevation and pedestrian walkways. UCSD includes two subsets: Ped1 and Ped2 which contain 7200 frames with 40 abnormal events and 2010 frames with 12 abnormal events, respectively. Videos are from the outdoor scene, recorded with a static camera at 10 fps. All other objects except for pedestrians are considered as irregularities.
 	
 	\item ShanghaiTech Campus dataset is considered to be one of the most comprehensive and realistic video anomaly detection data sets currently available. It contains 330 training videos and 107 test videos with 130 abnormal events on the campus of Shanghai University of Science and Technology. A total of 13 different scenarios and various types of anomalies are included. Due to the complexity of abnormal semantics, it is extremely challenging for current methods.
 	
	\item CUHK Avenue dataset is another representative dataset for video anomaly detection. It contains 16 and 21 video clips captured from a single camera for training and test, respectively. These videos were taken on Campus Avenue, a total of 30652 frames (15328 of training, 15324 of the test), mainly to detect pedestrians' abnormal movements, wrong movement directions, and the characteristics of abnormal objects.
	
	\item IITB-Corridor dataset is a large scale surveillance dataset with 483566 frames (301999 of training, 181567 of the test), which consists of group activities such as protest, chasing, fighting, sudden running as well as single person activities such as hiding face, loitering, unattended baggage, carrying a suspicious object and cycling (in a pedestrian area).
	\end{itemize}
	
  In order to facilitate comparison with other existing methods, we follow Morais et al.’s \cite{morais2019learning} strategy to manually filter out a set of video frames where the main anomalies are not related to humans, or where the individuals are not visible, or where objects can not be detected or tracked. The revised datasets are reserved as "Human-Related (HR) ShanghaiTech dataset" and "Human-Related (HR) Avenue dataset", respectively. Similarly, we adopt Jain et al.’s \cite{jain2021posecvae} strategy to filter IITB-Corridor to generate "Human-Related (HR) IITB-Corridor dataset"

  \subsubsection{Pose Tracking}
  
  The multi-person pose tracking algorithm is a crucial part of the framework. Firstly, the intensity of the input frame is normalized to the range of $[-1, 1]$, and each frame is adjusted to be with a resolution of $256 \times 256$ pixels. We use the object detection algorithm (YOLOv5x) \cite{ultralytics-yolov5} to locate each person. Next, the reID \cite{zheng2019joint} algorithm is used to extract the features of each pedestrian, including the clues such as body shape and clothing. Finally, the similarity score calculated by the bounding box coordinates from adjacent frames, and the score calculated by the reID algorithm is used together with the Hungarian \cite{kuhn1955hungarian} algorithm to track people. 
  
  The occlusion or sudden abnormal behavior of pedestrians cause the tracking ID of the human to be missing in some video frames. Existing algorithms assign a new tracking ID after the previously occluded person re-appears. We propose a “track-back” module that uses the similarity score calculated by the reID \cite{zheng2019joint} and motion vectors to judge whether to assign a new tracking ID or re-use original IDs to the identities in new frames. We use HRNet \cite{sun2019deep} to independently detect the skeleton in each video frame, by using the backbone of different sizes to balance the relationship between speed and accuracy, and explore the impact of accuracy on the effect of subsequent anomaly detection algorithms. Moreover, we utilize RAFT \cite{teed2020raft} to extract the optical flow of each frame as auxiliary information, which is suitable for low-resolution datasets.

  \subsubsection{Experimental Setting}
  
  We use a training batch size of 64 and the model was trained by using Stochastic Gradient Descent (SGD) with $\beta=0.9$, with 20, 60, 30 and 80 epochs on UCSD \cite{li2013anomaly}, ShanghaiTech \cite{luo2017revisit}, CUHK Avenue \cite{lu2013abnormal}, and IITB-Corridor \cite{rodrigues2020multi}, respectively. The initial learning rate is 0.5, the learning rate is adjusted according to a cosine annealing method \cite{loshchilov2016sgdr}. We train the model for one time in each scene, finally, different weight files are generated. All models are trained in an end-to-end manner using PyTorch \cite{paszke2017automatic} with an Nvidia GTX 2070.
  
 \subsection{Evaluation}

 \begin{table*}[ht]
 	\caption{Comparison of ROC AUC between HSTGCNN and other methods.} 
 	\label{tab:table1}
 	\begin{center}       
 		\begin{tabular*}{\hsize}{@{}@{\extracolsep{\fill}}ccccccccc@{}}
 			\toprule  
 			\midrule 
 			
 			Methods & UCSD ped1 & UCSD ped2 & HR-Avenue & Avenue & HR-ShanghaiTech & ShanghaiTech & HR-IITB-Corridor  & IITB-Corridor\\
 			\midrule[0.8pt]  
 			Frame-Pred \cite{liu2018future} & 83.10\%  & 95.40\% & - & 84.90\% & - & 72.80\% & - &64.65\%  \\
 			MPED-RNN \cite{morais2019learning}    & -  & - & 86.30\% & - & 75.40\% & 73.40\% & 68.07\% & 64.27\%   \\
 			w/ Mem \cite{park2020learning}     & -    & 97.00\%  & - & \textbf{88.50\%} & - & 70.50\% & - & -   \\
 			ST-GCAE \cite{markovitz2020graph}   & -    & -  & - & - & - & 71.60\% & - & -   \\
 			Multi-timescale \cite{rodrigues2020multi}   & - & - & 88.33\% & 82.85\% & 77.04\% & 76.03\% & - & 67.12\%   \\
 			PoseCVAE \cite{jain2021posecvae}   & -   & -  & 87.78\% & 82.10\% & 75.70\% & 74.90\% & 70.60\% & 67.34\%   \\
 			LSA \cite{abati2019latent}   & -    & 95.40\%  & - & - & - & 72.50\% & - & -   \\
 			Ano-Graph \cite{pourreza2021ano}   & -  & 96.68\%  & - & 86.20\% & - & 74.42\% & - & -   \\
 			AnomalyNet \cite{zhou2019anomalynet}   & \textbf{83.50\%}   & 94.90\%  & - & 86.10\% & - & - & - & -   \\
 			Normal Graph \cite{luo2021normal}   & - & - & - & 87.30\% & 76.50\% & 74.10\% & - & -   \\
 			\midrule  
 			HSTGCNN    & 83.39\%   & \textbf{97.73\%}  & \textbf{88.65\%} & 87.51\% & \textbf{83.40\% }& \textbf{81.80\%} & \textbf{73.92\%} & \textbf{70.46\%}   \\
 			
 			\midrule  
 			\bottomrule 
 			
 		\end{tabular*}
 	\end{center}
 \end{table*}

 \begin{table*}[ht]
 	\caption{Comparison of model parameters and inference speed between HSTGCNN and other methods (HR-ShanghaiTech dataset).} 
 	\label{tab:table2}
 	\begin{center}       
 		\begin{tabular*}{\hsize}{@{}@{\extracolsep{\fill}}ccc@{}}
 			\toprule 
 			\midrule
 			
 			Methods & Model parameters & Inference time (s)  \\
 			\midrule[0.8pt]  
 			Frame-Pred \cite{liu2018future}		& 7.7M  & 0.234  \\
 			YOLOv5x \cite{ultralytics-yolov5} + ST-GCAE \cite{markovitz2020graph}    & 87.8M + 793.4K  & 0.021 + 0.084   \\
 			YOLOv5x \cite{ultralytics-yolov5} + HRNet-w48 \cite{sun2019deep} + MPED-RNN \cite{morais2019learning}    & 87.8M + 63.6M + 25.46K    & 0.021 + 0.063 + 0.025   \\
 			\midrule  
 			YOLOv5x \cite{ultralytics-yolov5} + HRNet-w32 \cite{sun2019deep} + RAFT\cite{teed2020raft} + HSTGCNN          & 87.8M + 28.5M + 5.3M + 0.126K  & \textbf{0.021 + 0.029 + 0.019 + 0.0013}   \\
 			YOLOv5x \cite{ultralytics-yolov5} + HRNet-w48 \cite{sun2019deep} + RAFT\cite{teed2020raft} + HSTGCNN          & 87.8M + 63.6M + 5.3M + 0.126K  & 0.021 + 0.063 + 0.019 + 0.0013   \\
 			\midrule
 			\bottomrule 
 		 
 		\end{tabular*}
 	\end{center}
 \end{table*}
 \subsubsection{Comparisons with existing methods on accuracy}
 
 We train HSTGCNN on the training dataset and obtain weights for different scenarios. Fig.~\ref{fig4}. shows a curve describing the anomaly scores on different frames. Among them, for scenes with only normal behaviors, the anomaly score keeps low while for a frame with abnormal events, the anomaly score has a relatively high value.
 
 In the literature of anomaly detection, a popular evaluation method is to use the metrics of the Area Under Curve of the Receiver Operating Characteristic (ROC AUC). A higher ROC AUC value indicates better anomaly detection performance. In this paper, we leverage frame-level ROC AUC for performance evaluation following the work \cite{liu2018future}. 
 
\begin{figure} [ht]
 	\begin{center}
 		\begin{tabular}{c} 
 			\includegraphics[width=8.8cm]{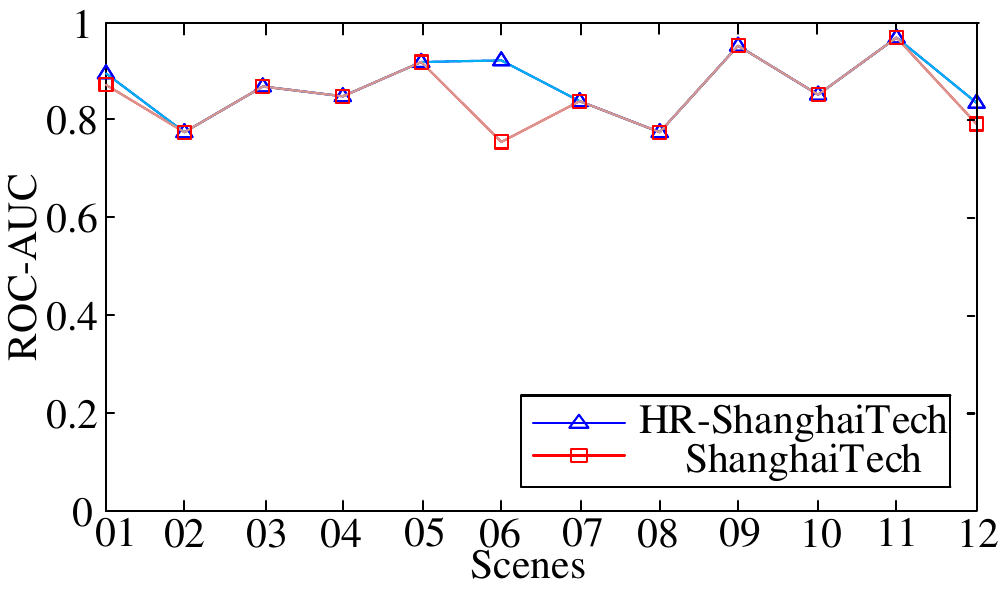}
 		\end{tabular}
 	\end{center}
 	\caption[example] 
 	{ \label{fig5} 
 		Independent ROC AUC for each scene in the HR-ShanghaiTech dataset and ShanghaiTech dataset}
 \end{figure}

 Table \ref{tab:table1} lists the comparison between the proposed HSTGCNN model and the latest state-of-the-art methods using ROC AUC. The ten methods are Frame-Pred \cite{liu2018future}, MPED-RNN \cite{morais2019learning}, w/Mem \cite{park2020learning}, ST-GCAE \cite{markovitz2020graph}, Multi-timescale \cite{rodrigues2020multi}, PoseCVAE \cite{jain2021posecvae}, LSA \cite{abati2019latent}, Ano-Graph \cite{pourreza2021ano}, AnomalyNet \cite{zhou2019anomalynet}, and Normal Graph \cite{luo2021normal}, some of them integrate a model focusing on appearance and motion with others dealing with the trajectories of human skeletons. From experiments, we can conclude that HSTGCNN outperforms the ten methods mentioned above on four public datasets, including Human-Related (HR) and original datasets. Although there are anomalies unrelated to humans in the video segment of the dataset, it still achieves the highest frame-level ROC AUC. As a result, compared with methods based on pixel reconstruction \cite{liu2018future},\cite{zhou2019anomalynet} and \cite{abati2019latent}, the model shows a stronger robustness. The noises of intensity-based features can be reduced by extracting the features describing human skeletons and motion instead of pixels. By levering a novel structure, our approach achieves advantages over RNNs and improves understanding of the global scenes in contrast to the coarse-grained STGCNN \cite{luo2021normal}, \cite{pourreza2021ano}. As expected, anomalous events about humans can be correctly detected in different scenes in the ShanghaiTech dataset, as shown in Fig.~\ref{fig5}. In addition, skeleton-based sequences and motion-related features are integrated by high-level graph representations to accommodate different resolution datasets. To conclude, effective structures, comprehensive feature representations involving high/low-level embeddings and adjustable weights according to scenes contribute to the significant improvement of the overall accuracy and robustness of the HSTGCNN method. 
 
 \subsubsection{Runtime}
 
 In terms of computational burdens, as shown in Table \ref{tab:table2}, the average detection time is measured for different network designs, and the model parameters and the amount of computation are compared. These models are implemented on Ubuntu systems with Nvidia GTX 2070 GPU. For the sake of fairness, the methods for comparison are with the same inputs and outputs. Our model for anomaly detection performs inference at a speed of 14.22 PFS and 9.59 FPS by using different sizes of HRNet for skeleton detection, respectively. The inference speed of other latest technologies ST-GCAE \cite{markovitz2020graph} is 9.52 FPS, Frame-Pred \cite{liu2018future} is 4.27 FPS and MPED-RNN \cite{morais2019learning} is 9.17 FPS. Compared with the four modules mentioned in Table \ref{tab:table2}, the runtime of tracking module can be ignored.
  
  As a result, the HSTGCNN network has a much higher efficiency than the above three methods. The design of the HSTGCNN network structure does not contain a large number of recursive structures (such as MPED-RNN \cite{morais2019learning}). At the same time, the number of learnable parameters in the proposed HSTGCNN is much less than in other models.

  \subsubsection{Qualitative results}
  
\begin{figure} [t]
  	\begin{center}
  		\begin{tabular}{c} 
  			\includegraphics[width=8.8cm]{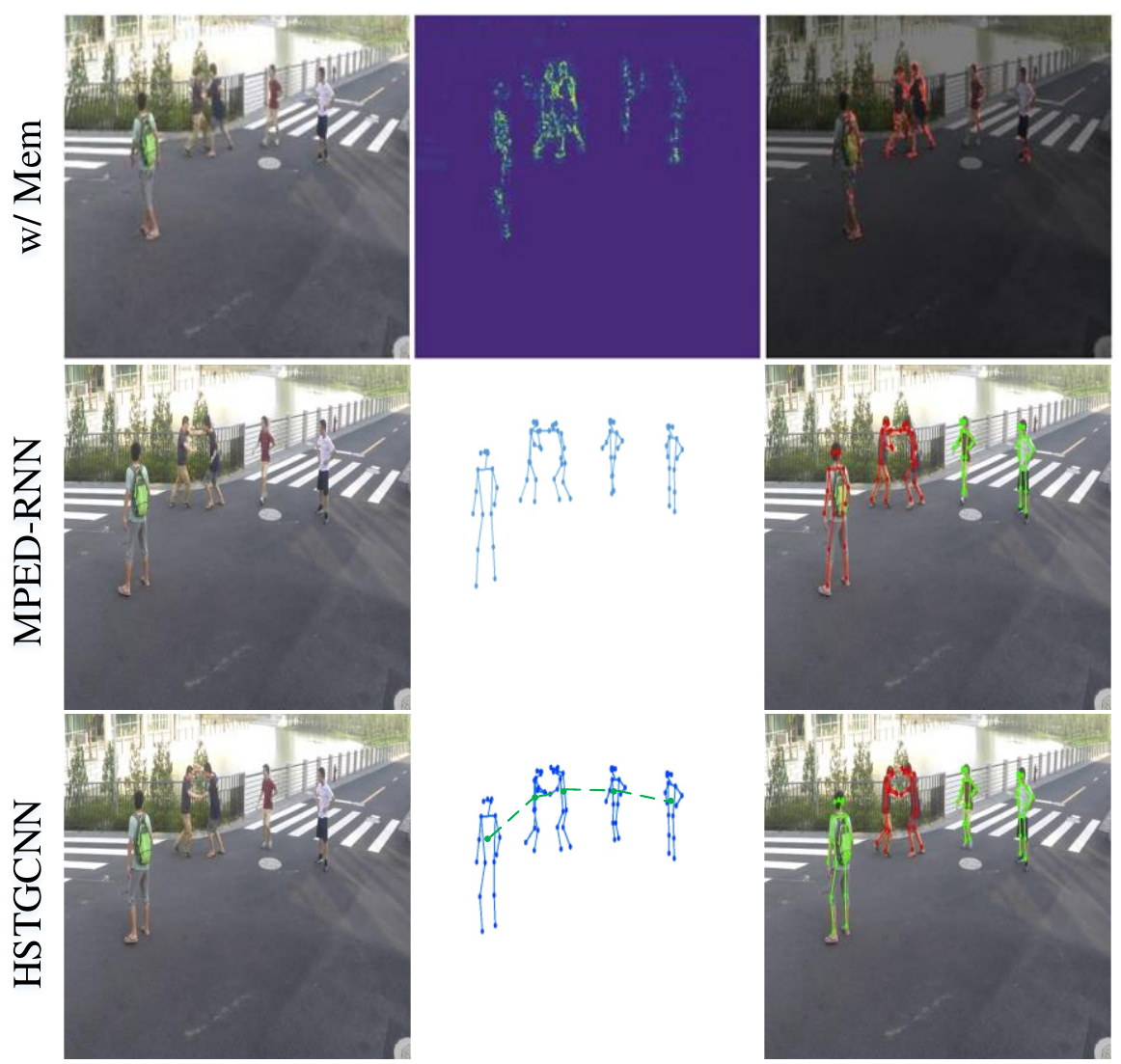}
  		\end{tabular}
  	\end{center}
  	\caption[example] 
  	{ \label{fig6} 
  		Qualitative results for future frame prediction of (top to bottom): w/ Mem \cite{park2020learning} model, MPED-RNN \cite{morais2019learning} model and our HSTGCNN model. Among them, input frames (left); prediction error (middle); abnormal regions (right). We can see that our model accurately localizes the regions with abnormal events.}
  \end{figure}

  Fig.~\ref{fig6}. shows the qualitative results of our model for future frame prediction and the other two latest methods on the ShanghaiTech dataset\cite{luo2017revisit}. The w/ Mem \cite{park2020learning} predicts anomalies based on reconstruction error. Although it can effectively detect abnormal regions, the result is noisy in that normal and abnormal regions are entangled. MPED-RNN \cite{morais2019learning} is an anomaly detection method based on the human skeleton. Even if it reduces the noise in the image and improves the detection accuracy, the anomaly regions are not accurately predicted in the video. Our hierarchical graphical inference method can more accurately detect abnormal regions in video frames, while also ensuring a higher frame-level detection accuracy.
  
  \subsection{Ablation study}
  
  \subsubsection{Influence of three branches combination}
  We analyze the network performance by altering the hyper-parameters of the outlier arbiter in the network. The outlier arbiter uses three prediction branches, $(L_{1}, L_{2}, L_{3})$, for hierarchical graphical inference, and these three branches represent three anomaly scores for each frame, the scores from different branches are weightedly summed to calculate the final anomaly score. In Section III-E, we explained in detail the meaning of the three prediction branches, which have a corresponding mapping relationship with high/low-level graph representations and affect the model detection strategy and provide different anomaly detection effects.
  
  \begin{table}[ht]
  	\caption{Influence of three branches combination.} 
  	\label{tab:table3}
  	\begin{center}       
  		\begin{tabular*}{\hsize}{@{}@{\extracolsep{\fill}}c|c|c|c|cc@{}}
  			\toprule
  			\midrule 
  			Methods & $L_{1}$      &   $L_{2}$    &  $L_{3}$  & HR-ShanghaiTech & HR-IITB-Corridor\\ 
  			\midrule[0.8pt]  
  			(a)     & $\surd$      &  $\times$    &  $\times$    & 68.60\%      & 68.97\%     \\
  			(b)     & $\times$     &  $\surd$     &  $\times$    & 81.20\%      & 67.52\%    \\
  			(c)     & $\times$     &  $\times$    &  $\surd$     & 78.00\%      & 69.75\%    \\
  			(d)     & $\surd$      &  $\surd$     &  $\times$    & 79.00\%      & 68.67\%    \\
  			(e)     & $\times$     &  $\surd$     &  $\surd$     & 81.70\%      & 69.98\%    \\
  			(f)     & $\surd$      &  $\times$    &  $\surd$     & 79.00\%      & 70.16\%    \\
  			(g)     & $\surd$      &  $\surd$     &  $\surd$     & \textbf{83.40\%}      & \textbf{73.92\%}    \\
  			\midrule
  			\bottomrule 
  		\end{tabular*}
  	\end{center}
  \end{table}

  Table \ref{tab:table3} shows the influence of hierarchical graph representations and combining branches. Experiments show that hierarchical graph representations are necessary, which covers more diverse abnormal events and improves the ability to handle abnormal semantics.

  \newcommand{\tabincell}[2]{\begin{tabular}{@{}#1@{}}#2\end{tabular}}  
   \begin{table}[t]
 	\caption{Influence of weight coefficients in different groups on accuracy (HR-ShanghaiTech dataset).} 
 	\label{tab:table4}
 	\begin{center}       
 		\begin{tabular*}{\hsize}{@{}@{\extracolsep{\fill}}cccc@{\centering}}

 			\toprule 
 			\midrule
 			Number of $K$ &    $W_{1}, W_{2}, W_{3}$   & Total loss during training   & ROC AUC \\
 			\midrule[0.8pt]  
 			3    & \tabincell{c}{$(0.2, 0.5, 0.3)$ \\ $(0.1, 0.8, 0.1)$\\ $(0.2, 0.7, 0.1)$}  & 41.2396   & 80.60\%      \\
 			\midrule  
 			6    & \tabincell{c}{$(0.3, 0.5, 0.2)$ \\ $(0.1, 0.5, 0.4)$\\ ……\\ $(0.6, 0.3, 0.1)$}       & 37.3425 & 82.20\%      \\ 
 			\midrule  
 			12   & \tabincell{c}{$(0.6, 0.1, 0.3)$ \\ $(0.1, 0.7, 0.2)$\\ ……\\ $(0.1, 0.5, 0.4)$\\ $(0.1, 0.1, 0.8)$}   & \textbf{34.6117}    & \textbf{83.40\%}      \\
 			\midrule
 			\bottomrule 
 		\end{tabular*}
 	\end{center}
 \end{table}
 
 \begin{figure*} [ht]
 	\begin{center}
 		\begin{tabular}{c} 
 			\includegraphics[width=18cm]{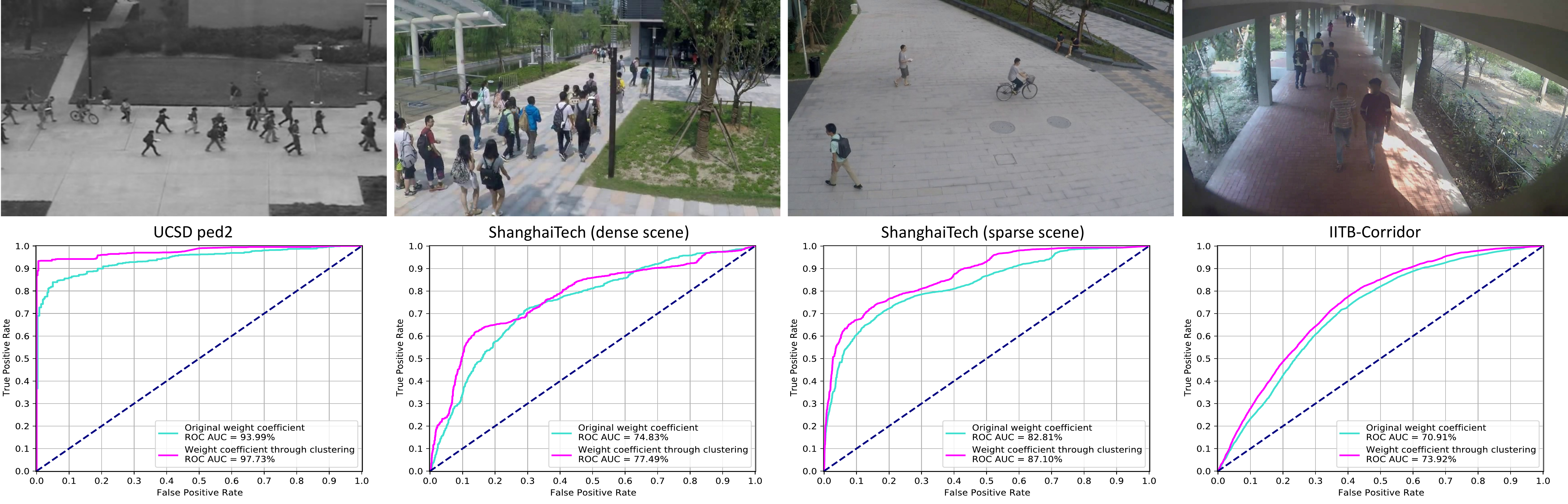}
 		\end{tabular}
 	\end{center}
 	\caption[example] 
 	{ \label{fig7} 
 		Comparing the original weight coefficient and the weight coefficient through clustering, the ROC AUC curve shows significant differences in the crowd or sparse scenarios on various resolutions datasets.}
 \end{figure*}
 \begin{table*}[!h]
   	\caption{ROC AUC performance of two methods with different pre-processing work.} 
   	\label{tab:table5}
   	\begin{center}     
   		 
   		\begin{tabular*}{\hsize}{@{}@{\extracolsep{\fill}}cccc@{}}
   			\toprule 
   			\midrule
   			Skeleton Trajectory Tracking & Anomaly Detection Modules &HR-ShanghaiTech & HR-IITB-Corridor  \\
   			\midrule[0.8pt]  
   			Morais et al.'s \cite{morais2019learning}  & MPED-RNN \cite{morais2019learning}   &   75.40\%    &   68.07\%\\
   			YOLOv5x \cite{ultralytics-yolov5} + HRNet-w32 \cite{sun2019deep}    & MPED-RNN \cite{morais2019learning}  &   74.47\%   &   66.73\%\\
   			YOLOv5x \cite{ultralytics-yolov5} + HRNet-w48 \cite{sun2019deep}    & MPED-RNN \cite{morais2019learning}  &   75.00\%   &   69.05\%\\
   			\midrule  
   			YOLOv5x \cite{ultralytics-yolov5} + HRNet-w32 \cite{sun2019deep}    & HSTGCNN     &   81.31\%   &   70.08\%\\
   			YOLOv5x \cite{ultralytics-yolov5} + HRNet-w48 \cite{sun2019deep}    & HSTGCNN     &   \textbf{83.40\%}   &   \textbf{73.92\%}\\
   			\midrule
   			\bottomrule 
   		\end{tabular*}
   	
   	\end{center}
   \end{table*}

  \subsubsection{Scene clustering and determination of weight coefficients}
   We divided training and test videos including 12 scenes into 3, 6, and 12 groups by clustering to explore the impact of the weight coefficients which are adapted to scene groups on the accuracy of the anomaly detection algorithm. Similar scenes correspond to the same group. In Section III-D, we clarified in detail the advantage of video clustering, which adjusts the weight coefficients of high/low level graph representations according to dense or sparse scenes, respectively, to enhance the robustness of the method.
      
   Table \ref{tab:table4} reports the influence of the weight coefficients in different groups on the accuracy of anomaly detection algorithm. In each group, we iterate repeatedly to select the best set of weight coefficients that minimize total output loss without labels in the training set. As can be seen, the lowest total loss and highest ROC AUC are usually achieved for $K$ = 12, that is, each scene corresponds to a set of weight coefficients and we use that value through all our experiments. 
      
   Fig.~\ref{fig7}. shows that with the usage of different strategies with different weights of branches under different scenarios among the datasets. Due to the size of pedestrians becomes smaller in dense scenes, the weight coefficient corresponding to high-level graph representations increases. On the contrary, the weight coefficient corresponding to low-level graph representations increases in sparse scenes including large pedestrians. The accuracy of our approach has been improved, which proves the weighted combinations contribute to an understanding of scenes.

  \subsubsection{Influence of using different tracking modules}
  In order to demonstrate the effectiveness of different components of the proposed framework, we divide the framework into a tracking module (including the different sizes of skeleton detection models) and a module for anomaly detection. We combine our tracking module with the anomaly detection module in Morais et al.'s \cite{morais2019learning} by using different sizes of HRNet \cite{sun2019deep} for skeleton detection. Finally, the above methods are compared on the HR-ShanghaiTech dataset \cite{luo2017revisit} and the HR-IITB-Corridor dataset \cite{rodrigues2020multi}. As shown in Table \ref{tab:table5}, the improvements in the accuracy of our method are contributed by larger HRNet for skeleton detection and our proposed HSTGCNN for anomaly detection.

  \subsection{Error Analysis}
  
  Although the proposed method based on human skeleton detection and tracking effectively improves the detection accuracy of abnormal events, it strongly depends on the performance of the model for detecting human skeletons. In some cases in the test set, the changes in the resolution or contrast of the picture cause the anomaly detection algorithm to fail. When multiple individuals are severely occluded by each other, the tracking algorithm produces wrong IDs, because the predictions on anomaly scores are negatively influenced. As a result, the problems that occur in pose estimation and tracking influence the performance of anomaly detection.

 \begin{figure} [ht]
 	\begin{center}
 		\begin{tabular}{c} 
 			\includegraphics[width=8.8cm]{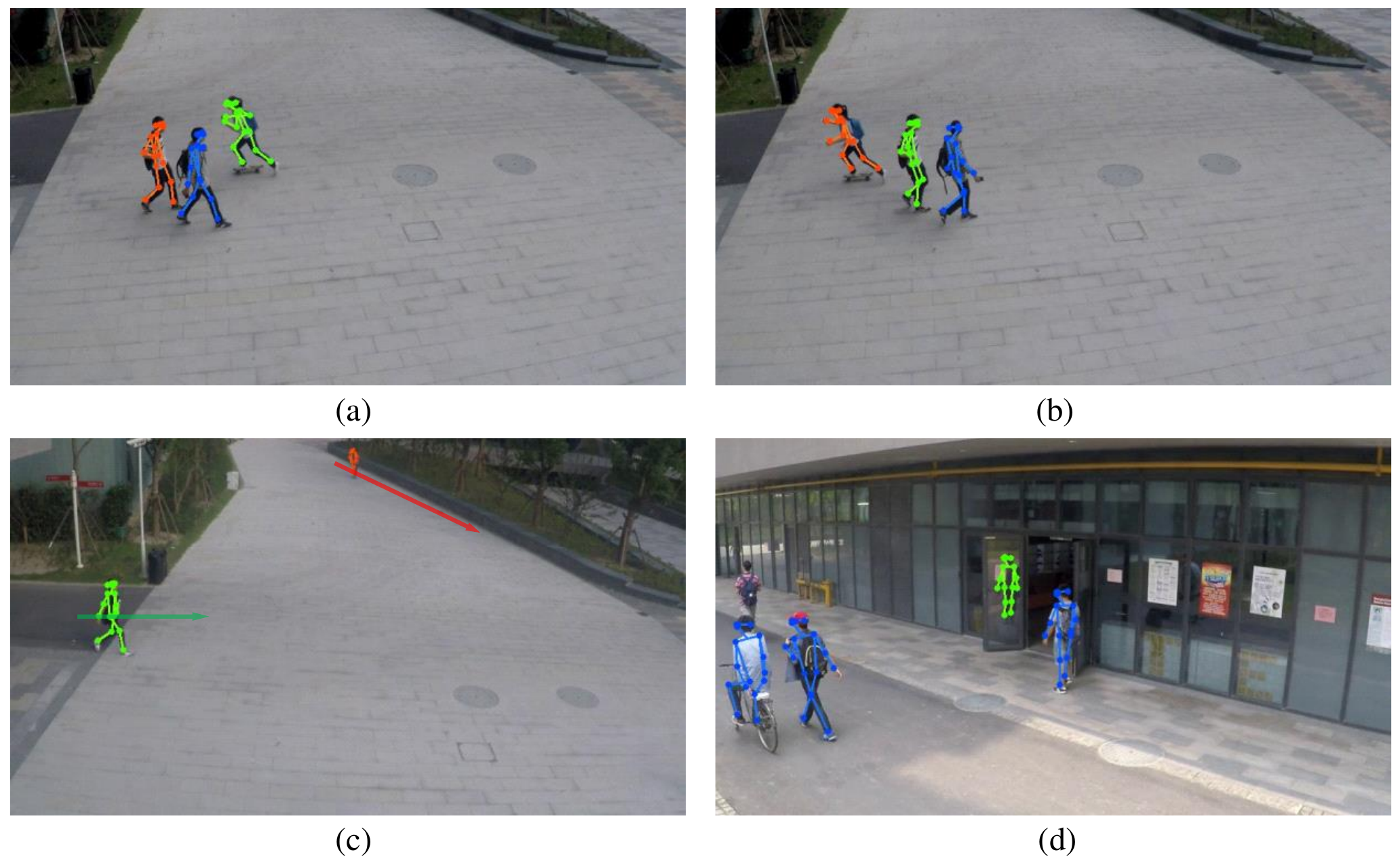}
 		\end{tabular}
 	\end{center}
 	\caption[example] 
 	{ \label{fig8} 
 		Some misrepresentations that may occur in pose tracking.}
 \end{figure}
 
 As shown in Fig.~\ref{fig8}.(a) \& (b), the pedestrian's exchange IDs due to occlusion. At the same time, the partial occlusion also leads to dramatic changes in poses which are judged as anomalies. In Fig.~\ref{fig8}.(c), due to the variances in scales, the moving speed of remote bicycles (small) is the same as that of a human in the vicinity (large), this influences the determination of anomalies based on moving speed. Even if we normalized the sizes of pedestrians, some false positives sometimes can not be avoided. In the case of the small-scale pedestrians of low resolutions, motion consistency among consecutive frames will be utilized for improvement. In Fig.~\ref{fig8}.(d), the algorithm erroneously detects the contour from the glass reflection, this introduces some interference for the subsequent tracking.
 
 \section{Conclusion}
 
 In this paper, we propose to use skeleton-based sequences and motion-related features to detect human-related anomalies without relying on any annotations about abnormal events. Specifically, we use a spatio-temporal convolutional network as our graphical feature extractor which is superior to other existing models in accuracy, memory consumption, and efficiency. The HSTGCNN we proposed integrates high-level graph representations with low-level graph representations. Low-level graph structure focuses on encoding the spatial and temporal embeddings of human body joints in high-resolution videos. High-level graph structure leverages the speed and directions of individuals and the interactions among multiple identities to describe abnormality in low-resolution videos. Meanwhile, high-level graph structure distinguishes the dense scenes including small people from sparse scenes including large ones by bounding boxes, skeletons, and optical flow. The weighted combinations of multiple branches that are better at handling different scenes achieve state-of-the-art performance. Additionally, a clustering method is leveraged to group scenes. Future research directions include the detection of human abnormal events with the assistance of semantic information and the improvement of pose tracking approaches.


\bibliographystyle{IEEEtran}
\bibliography{my_reference} 

\end{document}